\documentclass[runningheads]{llncs}
\usepackage{graphicx}
\usepackage{listings}
\usepackage{xcolor}
\usepackage [english]{babel}
\usepackage [autostyle, english = american]{csquotes}
\MakeOuterQuote{"}

\usepackage[frozencache=true,cachedir=minted-cache]{minted} 

\begin{document}

\title{Large Language Models in the Workplace: A Case Study on Prompt Engineering for Job Type Classification}

\titlerunning{Prompt Engineering for Job Classification}

\author{Benjamin Clavi\'{e}\inst{1} \and
Alexandru Ciceu\inst{2} \and
Frederick Naylor\inst{1} \and
Guillaume Souli\'{e}\inst{1} \and
Thomas Brightwell\inst{1}}

\authorrunning{B. Clavi\'{e} et al.}

\institute{Bright Network, Edinburgh, UK\\
\email{\{ben.clavie, firstname.lastname\}@brightnetwork.co.uk}\\
\and
Silicon Grove, Edinburgh, UK\\
\email{alex@silicongrove.co}}
\maketitle  
\begin{abstract}

This case study investigates the task of job classification in a real-world setting, where the goal is to determine whether an English-language is appropriate for a graduate or entry-level position. We explore multiple approaches to text classification, including supervised approaches such as traditional models like Support Vector Machines (SVMs) and state-of-the-art deep learning methods such as \textsc{DeBERTa}. We compare them with Large Language Models (LLMs) used in both few-shot and zero-shot classification settings. To accomplish this task, we employ prompt engineering, a technique that involves designing prompts to guide the LLMs towards the desired output. Specifically, we evaluate the performance of two commercially available state-of-the-art GPT-3.5-based language models, \textsc{text-davinci-003} and \textsc{gpt-3.5-turbo}. 
We also conduct a detailed analysis of the impact of different aspects of prompt engineering on the model's performance.\\
Our results show that, with a well-designed prompt, a zero-shot \textsc{gpt-3.5-turbo}classifier outperforms all other models, achieving a 6\% increase in Precision@95\% Recall compared to the best supervised approach. Furthermore, we observe that the wording of the prompt is a critical factor in eliciting the appropriate "reasoning" in the model, and that seemingly minor aspects of the prompt significantly affect the model's performance. 

\end{abstract}

\section{Introduction}

The combination of broadened access to higher education and rapid technological advancement with the mass-adoption of computing has resulted in a number of phenomena. The need for computational tools to support the delivery of quality education at scale has been frequently highlighted, even allowing for the development of an active academic subfield\cite{las}. At the other end of the pipeline, technological advances have caused massive changes in the skills required for a large amount of jobs\cite{wefreport}, with some researchers also highlighting a potential mismatch between these required sets of skills and the skills possessed by the workforce\cite{skillmismatch}. These issues lead to a phenomenon known as the "education-job mismatch", which can lead to negative effects on lifetime income\cite{veselinovic2020effect}. \\
Due in part to these factors, the modern employment landscape can be difficult to enter for recent graduates, with recent LinkedIn surveys showing that over a third of "entry-level" positions require multiple years of experience, and more than half of such positions requiring 3 years experience in certain fields or extremely specific skills\cite{linkedin}. As a result, it has been noted that entering the job market is an increasingly difficult task, now demanding considerable time and effort\cite{bbcjobs}.
While computational advances are now commonly used to support education and to assist workers in their everyday work, there is a lack of similarly mature technological solutions to alleviate the issues presented by exiting education to enter the workplace. We believe that the rapid development of machine learning presents a powerful opportunity to help ease this transition. \\
The case study at the core of this paper focuses on one of the important tasks to build towards this objective: Graduate Job Classification. Given a job posting containing its title and description, our aim is to be able to automatically identify whether or not the job is a position fit for a recent graduate or not, either because it requires considerable experience or because it doesn't require a higher education qualification. In light of the information presented above, as well as the sheer volume of job postings created every day, this classification offers an important curation. This would allow graduates to focus their efforts on relevant positions, rather than spending a considerable amount of time filtering through large volumes of jobs, which is non-trivial due to often obfuscated requirements.\cite{linkedin} As a point of reference, the number of total job postings in the United Kingdom alone in the July-September 2022 period exceeded 1.2 million\cite{onsjobs}. \\
This task contains a variety of challenges, the key one being the extreme importance of minimizing false negatives, as any false negative would remove a potentially suitable job from a job-seeker's consideration when the list is presented to them. On the other hand, with such large volumes of posting, too many false positives would lead to the curated list being too noisy to provide useful assistance. A second major challenge is the reliance of the task on subtle language understanding, as the signals of a job's suitability can be very weak. \\
In this paper, we will evaluate a variety of text classification approaches applied to the English-language Graduate Job Classification task. In doing so, we will (i) show that the most recent Large Language Models (LLMs), based on Instruction-Tuned GPT-3\cite{gpt3,instructgpt}, can leverage the vast wealth of information acquired during their training to outperform state-of-the-art supervised classification approaches on this task and that (ii) proper prompt engineering has an enormous impact on LLM downstream performance on this task, contributing a real-world application to the very active research on the topic of prompt engineering\cite{promptcatalog,ppp}.

\section{Background}

Since the introduction of the Transformer architecture\cite{transformers} and the rise of transfer learning to leverage language models on downstream tasks\cite{ulmfit,elmo}, the field of NLP has undergone rapid changes. Large pre-trained models such as BERT\cite{bert} and later improvements, like DeBERTa\cite{deberta}, have resulted in significant performance improvements, surpassing prior word representation methods such as word vectors\cite{word2vec}. The development of libraries such as HuggingFace Transformers\cite{huggingface} has further contributed to making these models ubiquitous in NLP applications.

These advances resulted in a paradigm shift in NLP, focusing on the use or fine-tuning of extremely large, generalist, so-called "foundation models" rather than the training of task-specific models\cite{foundation}. This resulted in the frequent occurrence of \textit{paradigm shift}, where researchers focused on ways to reframe complex tasks into a format that could fit into tasks where such models are known to be strong, such as question-answering or text classification\cite{paradigmshift}.

In parallel to these fine-tuning approaches, there has been considerable work spent on the development of generative Large Language Models (LLMs), whose training focuses on causal generation: the task of predicting the next token given a context\cite{gpt2}. The release of GPT-3 showed that these models, on top of their ability to generate believable text, are also few-shot learners: given few examples, they are capable of performing a variety of tasks, such as question answering\cite{gpt3}.

Going further, very recent developments have shown that LLMs can reach noticeably better performance on downstream applications through \textbf{instruction-tuning}: being fine-tuned to specifically follow natural language instructions to reach state-of-the-art performance on many language understanding tasks\cite{instructgpt}.

LLMs, being trained on billions of tokens, have been shown to be able to leverage the vast amount of knowledge found in their training data on various tasks, with performance increasing via both an increase in model and training data size, following complicated scaling laws\cite{gopher}. This has paved the way for the appearance of a new approach to NLP applications, focusing on exploring ways to optimally use this large amassed knowledge: \textbf{prompt engineering}\cite{ppp}.

Prompt Engineering represents a new way of interacting with models, through natural language queries. It has gathered considerable research attention in the last year. Certain ways of prompting LLMs, such as Chain-of-Thought (CoT) prompting, have been shown to be able to prompt reasoning which considerably improves the models' downstream performance\cite{cot}. Additional research has showcased ways to bypass certain model weaknesses. Notably, while LLMs are prone to mathematical errors, they are able to generate executable Python code to compute the requested results through specific prompting\cite{PAL}.

Other efforts have showcased reasoning improvements by relying on a model self-verifying its own reasoning in a subsequent prompt, which improves performance\cite{zeroshot}. All these approaches have shown that LLMs can match or outperform state-of-the-art results on certain tasks, while requiring little to no fine-tuning.

\section{Experimental Setup}
\subsection{Data and Evaluation}

\begin{table}
\caption{High-level description of the data used to train and evaluate models.}\label{data}
\begin{center}
\begin{tabular}{|l|c|c|c|c|}
\hline
          & \textbf{Example \#} & \textbf{Proportion} & \textbf{\begin{tabular}[c]{@{}c@{}}Median\\ Token \#\end{tabular}} & \textbf{\begin{tabular}[c]{@{}c@{}}Token \#\\ standard dev.\end{tabular}} \\ \hline
\textbf{GRAD}         & 3082    & 30.8\%  & 809    & 338           \\ \hline
\textbf{NON\_GRAD}    & 6918    & 69.2\%  & 831    & 434           \\ \hline
\textbf{Full Dataset} & 10000   & 100\%   & 821    & 389           \\ \hline
\end{tabular}
\end{center}
\end{table}
\textbf{Data}
Our target task is Graduate Job Classification. It is a binary classification, where, given a job posting containing both the job title and its description, the model must identify whether or not the job is a position fit for a recent graduate or not, either because it requires more experience or doesn't require higher education. In practice, over 25,000 jobs are received on a daily basis, with fewer than 5\% of those appropriate for graduates.

Curating positions fit for recent graduates is extremely time-consuming and is therefore one of the areas where technological advances can help simplify the process of entering the workplace. In practice, over 20,000 jobs go through our deployed model on a daily basis, with fewer than 5\% of those appropriate for graduates.

Our data is gathered from a large selection of UK-based jobs over a period of two years. These jobs were manually filtered into "Graduate" and "Non-Graduate" categories by human annotators working for Bright Network. All annotators work as part of a team dedicated to ensuring the quality of jobs and follow predefined sets of guidelines. Guidelines are frequently reviewed by domain experts, and feedback on annotation quality is gathered on a weekly basis. This is our \textit{silver} dataset. Unlike our gold standard described below, sampled from it and iterated upon, this is a single-pass annotation process, and individual mistakes can occasionally be present.

The gold standard dataset used in this study is a subset of the original data, containing job postings whose original label was further reviewed manually. Only jobs where inter-annotator agreement was reached were kept, until reaching a data size of 10,000. We use the label \textsc{GRAD} for jobs suitable for graduates and \textsc{NON\_GRAD} for all other jobs.

Before being used as model input, all job descriptions are prepended by the posting's title. A general description of the data is presented in Table~\ref{data}, including the distribution of labels and information about the token counts within documents. Overall, the median length of both \textsc{GRAD} and \textsc{NON-GRAD} jobs is similar, and the final dataset is made up of roughly 30\% \textsc{GRAD} jobs and 70\% \textsc{NON-GRAD} jobs.

\textbf{Evaluation}
We use the Precision at 95\% Recall (P@95\%R) for the \textsc{GRAD} label as our main metric. This means that our primary method of evaluation is the Precision (the measure of how good the model is at avoiding false positives), obtained by the model while maintaining a Recall of at least 95\%, which means the model detects at least 95\% of positive examples.
We chose this metric as the classifier cannot be deployed in production with a low recall, as it is extremely damaging to remove suitable jobs from graduates' consideration. Our goal is to ensure that Recall remains above a specific threshold while achieving the best possible precision at this threshold and help process the tens of thousands of jobs received daily. We also report the P@85\%R, to give a better overview of the models' performance. 
To facilitate LLM evaluation, we split our data into stratified training and test sets, respectively containing 7000 (70\%) and 3000 (30\%) examples, rather than using cross-validation. 

\subsection{Baselines}

\textbf{Keyword}  We report the results for a simple, keyword and regular expression approaches to the task. We, along with our annotators, built a list of common phrases and regular expressions indicating that a job is suitable for a recent graduate. We then perform a simple look-up within the postings, which gives us a lower bound for performance. An example of such an approach would be matching the words "Graduate" and "Junior" in job titles, or looking for strings such as "{is|would be} suitable for {graduate|student}" within the posting itself.

\textbf{SVM} We present the results of a non-deep learning baseline method, which involves using a Support Vector Machine (SVM) classifier with a tf-idf text representation, which has been shown to produce robust baseline results, even reaching state-of-the-art results in some domain-specific tasks\cite{bensvm}.
\subsection{Supervised Classifiers}

\textbf{ULMFiT} We report the results for ULMFiT, an RNN-based approach to training a small language model before fine-tuning it for classification\cite{ulmfit}. We pre-train the ULMFiT language model on an unlabeled dataset of 50000 job postings, before fine-tuning the classifier on the data described above.

\textbf{DeBERTa-V3} We fine-tune a DeBERTa-V3-Base model, a refined version of DeBERTa\cite{deberta} and which achieves state-of-the-art performance on a variety of text classification tasks\cite{debertav3}. We follow the method used in the paper introducing the model, with a maximum sequence length of 512. For any longer document, we report results using the first 100 tokens and the trailing 412 tokens of the document. This approach yielding the best results is likely due to most job descriptions frequently outlining the position's requirements towards the end.

\subsection{Large Language Models}

We use a temperature of 0 for all language models. The temperature controls the degree of randomness applied to the tokens outputted by the language model. A temperature of 0 ensures the sampling favors the highest probability token in all cases, resulting in a deterministic output.

\textbf{GPT-3.5 (text-davinci-002\&text-davinci-003)} We report our results on two variants of GPT-3\cite{gpt3}\footnote{These models are accessed through OpenAI's API.}. These models are LLMs further trained to improve their ability to follow natural language instructions\cite{instructgpt}. Although the detailed differences between the two models are not made public, \textsc{davinci-003} is a refinement of \textsc{davinci-002}, better at following instructions\footnote{Introduced by OpenAI in a blog post rather than a technical report: https://help.openai.com/en/articles/6779149-how-do-text-davinci-002-and-text-davinci-003-differ}.

\textbf{GPT-3.5-turbo (gpt-3.5-turbo-0301)} We evaluate GPT-3.5-turbo\footnote{This model is also accessed through OpenAI's API.}, a model optimized for chat-like interactions\cite{gptch}. To do so, we modified all our prompts to fit the conversation-like inputs expected by the model. GPT-3.5-turbo is the focus of our prompt engineering exploration.

\section{Overall Results}

In Table~\ref{results}, we report the P@95R\% and P@85R\% for all models evaluated. We report a score of 0 if the model is unable to reach the recall threshold. For all approaches for which we do not have a way to target a specific Recall value, we also provide their Recall metric.

\begin{table}[]
\caption{Results for all evaluated models.}\label{results}
\begin{tabular}{|l|c|c|c|c|c|c|c|c|}
\hline
        & Keyword & SVM  & ULMFiT & DeBERTaV3       & davinci-002 & davinci-003 & gpt-3.5       \\ \hline
P@95\%R & 0       & 63.1 & 70.2   & 79.7         & 0        & 80.4        & \textbf{86.9} \\ \hline
P@85\%R & 0       & 75.4 & 83.2   & \textbf{89.0}     & 72.6        & 80.4        & 86.9          \\ \hline
Recall & 80.2       & N/A & N/A   & N/A & 72.2         & 95.6        & 97         \\ \hline
\end{tabular}
\end{table}

Overall, we notice that SVMs, as often, are a strong baseline, although they are outperformed by both supervised deep learning approaches. However, they are outperformed by both of our supervised approaches. DeBERTaV3 achieves the highest P@85\%R of all the models, but is beaten by both \textsc{davinci-003} and \textsc{GPT-3.5} on the P@95\%R metric, which is key to high-quality job curation.

We notice overall strong performance from the most powerful LLMs evaluated, although \textsc{davinci-002} fails to reach our 95\% Recall threshold and trails behind both ULMFiT and DeBERTaV3 at an 85\% recall threshold. On the other hand, \textsc{davinci-003} outperforms DeBERTaV3, while \textbf{\textsc{GPT-3.5}} is by far the best-performing model on the P@95\%R metric, with a 7.2 percentage point increase.

Overall, these results show that while our best-performing supervised approach obtains better metrics at lower recall thresholds, it falls noticeably behind LLMs when aiming for a very low false negative rate.

\section{LLMs \& Prompt Engineering}

\begin{table}[]
\caption{Overview of the various prompt modifications explored in this study.}\label{promptoverview}
\begin{tabular}{l|l}
Short name & Description                                                                \\ \hline
Baseline   & Provide a a job posting and asking if it is fit for a graduate.            \\
CoT        & Give a few examples of accurate classification before querying.            \\
Zero-CoT   & Ask the model to reason step-by-step before providing its answer.          \\
rawinst    & Give instructions about its role and the task by adding to the user msg.   \\
sysinst    & Give instructions about its role and the task as a system msg.             \\
bothinst   & Split instructions with role as a system msg and task as a user msg.       \\
mock       & Give task instructions by mocking a discussion where it acknowledges them. \\
reit       & Reinforce key elements in the instructions by repeating them.              \\
strict     & Ask the model to answer by strictly following a given template.            \\
loose      & Ask for just the final answer to be given following a given template.      \\
right      & Asking the model to reach the right conclusion.                            \\
info       & Provide additional information to address common reasoning failures.       \\
name       & Give the model a name by which we refer to it in conversation.             \\
pos        & Provide the model with positive feedback before querying it.              
\end{tabular}
\end{table}

In this section, we will discuss the prompt engineering steps taken to reach the best-performing version of \textsc{GPT-3.5}. We will largely focus on its chat-like input, although similar steps were used for other language models, minus the conversational format. Apart from the use of system messages, we noticed no major differences in prompt impact between models.

For each modification, we will provide an explanation of the changes, or, where relevant, a snippet highlighting the modification. An overview of all prompt modifications used is presented in Table~\ref{promptoverview}. We evaluate the impact of each change on the model's performance, but also on its ability to provide its answer in a specified format rather than as free text, which we call \textbf{Template Stickiness}.

Our approach to prompt engineering, as described in this section, follows the ChatML \cite{chatml} prompting format used by OpenAI for their GPT family of models. To help readability, we do not directly reproduce the XML-like or JSON format used by ChatML but provide simple \textsc{variable-like} identifiers for our prompt modifications in the examples below.
\subsection{Eliciting Reasoning}

\textbf{Zero-Shot Prompting}
We set our baseline by simply prompting the model with our question with no further attempt to induce reasoning ('\textbf{Baseline}'):

\begin{minted}[breaklines,breakanywhere,bgcolor=lightgray]{text}
For the given job:
{job_posting}
---------
Is this job (A) a job fit for a recent graduate,
or (B) a job requiring more professional experience.
Answer:
\end{minted}

\textbf{Few-shot CoT}
We then experiment with few-shot chain-of-thought prompting\cite{cot}, by providing the model with successful classification examples. We do so using the \textsc{gpt-3.5} chat format, mocking a conversation between the user, and the assistant, who elaborates on his reasoning before answering with (A) or (B). We prepend our query by providing the model with two examples\footnote{Due to the long token length of job postings, providing it with more than two examples required us to truncate the postings, which resulted in a degradation in performance} (\textbf{'CoT'}). We do so in the folloiwng format:
\begin{minted}[breaklines,breakanywhere,bgcolor=lightgray, escapeinside=||]{text}
|\textcolor{red}{user\_message\_1 = """For the given job:}|
|\textcolor{red}{\{job\_posting\}}|
|\textcolor{red}{---------}|
|\textcolor{red}{Is this job (A) a job fit for a recent graduate, or (B) a job requiring more professional experience."""}|

|\textcolor{red}{assistant\_message\_1 = "This job appears to be a senior position, as it mentions requiring experience interacting }||\textcolor{red}{with C-level stakeholder in intense environments and [...]. Therefore, this is (B) a job requiring more professional experience}|
|\textcolor{red}{user\_message\_2 =}| [...]
\end{minted}

\textbf{Zero-shot CoT}
We then attempt to elicit reasoning without providing the model any example, through Zero-shot Chain-of-Thought\cite{zeroshot} (\textbf{'Zero-CoT'}). We expect that this approach will perform well, as job postings are found in large quantity in data used to train the model, and identifying whether a job is fit for a graduate does not require expert domain knowledge. We attempt to elicit reasoning by prompting the model think step-by-step, as follows:

\begin{minted}[breaklines,breakanywhere,bgcolor=lightgray, escapeinside=||]{text}
For the given job:
{job_posting}
---------
Is this job (A) a job fit for a recent graduate,
or (B) a job requiring more professional experience.
Answer: |\textcolor{red}{Let's think step by step,}|
\end{minted}

\subsection{Initial Instructions}

We then explore the impact of providing the model with instructions describing both its role and task. A notable difference between textsc{davinci-003} and the \textsc{gpt-3.5} chat format is that the latter introduces a new aspect to prompt engineering, which was not found in previous ways to interact with language models: the ability to provide a \textit{system} message to the system. We explore multiple ways of providing instructions using this system message. 

\textbf{Giving Instructions} We provide information to the model about its role role as well as a description of its task:
\begin{minted}[breaklines,breakanywhere,bgcolor=lightgray, escapeinside=||]{text}
|\textcolor{teal}{role}| = """You are an AI expert in career advice. You are tasked with sorting through jobs by analysing their content and deciding whether they would be a good fit for a recent graduate or not."""
|\textcolor{teal}{task}| = """A job is fit for a graduate if it's a junior-level position that does not require extensive prior professional experience. I will give you a job posting and you will analyse it, to know whether or not it describes a position fit for a graduate."""
\end{minted}

\textbf{Instructions as a user or system message} There is no clear optimal way to use the \textit{system} prompt, as opposed to passing instructions as a \textit{user} query. The \textbf{'rawinst'} approach, explained above, passes the whole instructions to the model as a user query. We evaluate the impact of passing the whole instructions as a system query (\textbf{'sysinst'}), as well as splitting them in two, with the model's role definition passed as a system query and the task as a user query (\textbf{bothinst}).

\textbf{Mocked-exchange instructions}
We attempt to further take advantage of the LLM's fine-tuned ability to follow a conversational format by breaking down our instructions further (\textbf{'mock'}). We iterate on the \textbf{bothinst} instruction format, by adding an extra confirmation message from the model:
\begin{minted}[breaklines,breakanywhere,bgcolor=lightgray, escapeinside=||]{text}
user_message_1 = """A job is fit for a graduate [...] |\textcolor{red}{Got it?"""}|
|\textcolor{red}{assistant\_message\_1 = "Yes, I understand. I am ready to analyse your job posting."}|
\end{minted}

\textbf{Re-iterating instructions}
We further modify the instructions by introducing a practice commonly informally discussed but with little basis: re-iterating certain instructions (\textbf{'reit'}). In our case, this is done by appending a reminder to the system message, to reinforce the perceived expertise of the model as well as the importance of thinking step-by-step in the task description:
\begin{minted}[breaklines,breakanywhere,bgcolor=lightgray, escapeinside=||]{text}
|\textcolor{teal}{system\_prompt}| ="""You are an AI expert in career advice. You are tasked with sorting through jobs by analysing their content and deciding whether they would be a good fit for a recent graduate or not. |\textcolor{red}{Remember, you're the best AI careers expert and will use your expertise to provide the best possible analysis}|"""
|\textcolor{teal}{user\_message\_1}| = """[...] I will give you a job posting and you will analyse it, |\textcolor{red}{step-by-step}|, to know whether [...]"""
\end{minted}

\subsection{Wording the prompt}

\textbf{Answer template}
We experiment with asking the model to answer by following specific templates, either requiring that the final answer (\textbf{'loose'}), or the full reasoning (\textbf{'strict'}) must adhere to a specific template. We experiment with different wordings for the template, with the best-performing ones as follows:
\begin{minted}[breaklines,breakanywhere,bgcolor=lightgray, escapeinside=||]{text}
|\textcolor{teal}{loose}| = """[...]|\textcolor{red}{Your answer must end with:}|
|\textcolor{red}{Final Answer: This is a}| (A) job fit for a recent graduate or
a student OR (B) a job requiring more professional experience.
Answer: Let's think step-by-step,"""
|\textcolor{teal}{strict}| = """[...]|\textcolor{red}{You will answer following this template:}|
|\textcolor{red}{Reasoning step 1:}|\n|\textcolor{red}{Reasoning step 2:}|\n|\textcolor{red}{Reasoning step 3:}|\n
|\textcolor{red}{Final Answer: This is a}| (A) job fit for a recent graduate or
a student OR (B) a job requiring more professional experience.
Answer: |\textcolor{red}{Reasoning Step 1:}|"""
\end{minted}

\textbf{{The right conclusion}} We evaluate another small modification to the prompt to provide further positive re-inforcement to the model: we ask it reason in order to reach the right conclusion, by slightly modifying our final query:

\begin{minted}[breaklines,breakanywhere,bgcolor=lightgray, escapeinside=||]{text}
Answer: Let's think step-by-step |\textcolor{red}{to reach the right conclusion,}|
\end{minted}

\textbf{Addressing reasoning gaps} While analysing our early results, we noticed that the model can misinterpret instructions given to it, and produce flawed reasoning as a result. This manifested in attempts to over-generalise:
\begin{quote}
\textit{This job requires experience, but states that it can have been acquired through internships. However, not all graduates will have undergone internships. Therefore, (B) this job is not fit for all graduates.}
\end{quote}

We attempt to alleviate this by providing additional information in the model's instruction: 

\begin{minted}[breaklines,breakanywhere,bgcolor=lightgray, escapeinside=||]{text}
task = "A job is fit for a graduate if it's a junior-level position that does not require extensive prior professional experience. |\textcolor{red}{When analysing the experience required, take into account that requiring internships is still fit for a graduate}|. I will give you a job [...]
\end{minted}

\subsection{The importance of subtle tweaks}

\textbf{Naming the Assistant}
A somewhat common practice, as shown by Microsoft code-naming its Bing chatbot "Sydney" \footnote{As demonstrated by the widely circulated prompt https://simonwillison.net/2023/Feb/15/bing/}, is to give LLMs a nickname by which they can be referred to. We modified our initial system prompt, as well as the user mocked-instructions, to refer to our model as Frederick (\textbf{'name'}), as follows:

\begin{minted}[breaklines,breakanywhere,bgcolor=lightgray,escapeinside=||]{text}
|\textcolor{teal}{system\_prompt}| = "|\textcolor{red}{You are Frederick,}| an AI expert
in career advice. [...]"
[...]
|\textcolor{teal}{first\_assistant\_response}| = "Yes, I understand. |\textcolor{red}{I am Frederick,}|
and I will analyse your job posting."
\end{minted}

We tested multiple other names, chosen randomly from a list of common first-names in English-speaking countries. We noticed no significant variation in performance no matter the name given, and all resulted in a similar imporvement.

\textbf{Positive Feedback}
It has been anecdotally noted that giving positive reinforcement to \textsc{gpt-3.5} can lead to better performance on some tasks \footnote{As reported by OpenAI, a partnered developer found that positive reinforcement resulted in increased accuracy.}. We thus prepend our main prompt with a positive reaction to the model's mocked acknowledgement of our instructions (\textbf{'pos'}):
\begin{minted}[breaklines,breakanywhere,bgcolor=lightgray,escapeinside=||]{text}
|\textcolor{red}{Great! Let's begin then :)}|
For the given job: [...]
\end{minted}
\subsection{Best-Performing Final Prompt}

\begin{minted}[breaklines,breakanywhere,bgcolor=lightgray,escapeinside=||]{text}
|\textcolor{teal}{system}| = "You are Frederick, an AI expert in career advice. You
are tasked with sorting through jobs by analysing their content
and deciding whether they would be a good fit for a recent
graduate or not.",
|\textcolor{teal}{user\_1}| =  """A job is fit for a graduate if it's a
junior-level position that does not require extensive prior
professional experience.
When analysing the experience required, take into account that
requiring internships is still fit for a graduate. I will
give you a job posting and you will analyse it, step-by-step,
to know whether or not it describes a position fit for a
graduate. Got it?"""
|\textcolor{teal}{assistant\_1}| = "Yes, I understand. I am Frederick, and I will
analyse your job posting.",
|\textcolor{teal}{user\_2}| = """Great! Let's begin then :)
For the given job:
{job_posting}
---------
Is this job (A) a job fit for a recent graduate, or
(B) a job requiring more professional experience.
Answer: Let's think step by step to reach the right
conclusion"""
\end{minted}

\section{Prompt Engineering Results and Discussion}
\begin{table}[]
\caption{Impact of the various prompt modifications.}\label{promptresults}
\begin{tabular}{l|cccc}
                                        & Precision     & Recall        & F1            & Template Stickiness    \\ \hline
\textit{Baseline}                       & \textit{61.2} & \textit{70.6} & \textit{65.6} & \textit{79\%}          \\ \hline
\textit{CoT}                            & \textit{72.6} & \textit{85.1} & \textit{78.4} & \textit{87\%}          \\ \hline
\textit{Zero-CoT}                       & \textit{75.5} & \textit{88.3} & \textit{81.4} & \textit{65\%}          \\
\textit{+rawinst}                       & \textit{80}   & \textit{92.4} & \textit{85.8} & \textit{68\%}          \\
\textit{+sysinst}                       & \textit{77.7} & \textit{90.9} & \textit{83.8} & \textit{69\%}          \\
\textit{+bothinst}                      & \textit{81.9} & \textit{93.9} & \textit{87.5} & \textit{71\%}          \\ \hline
+bothinst+mock                          & 83.3          & 95.1          & 88.8          & 74\%                   \\
+bothinst+mock+reit                     & 83.8          & 95.5          & 89.3          & 75\%                   \\ \hline
\textit{+bothinst+mock+reit+strict}     & \textit{79.9} & \textit{93.7} & \textit{86.3} & \textit{\textbf{98\%}} \\
\textit{+bothinst+mock+reit+loose}      & \textit{80.5} & \textit{94.8} & \textit{87.1} & \textit{95\%}          \\ \hline
+bothinst+mock+reit+right               & 84            & 95.9          & 89.6          & 77\%                   \\
+bothinst+mock+reit+right+info          & 84.9          & 96.5          & 90.3          & 77\%                   \\
+bothinst+mock+reit+right+info+name     & 85.7          & 96.8          & 90.9          & 79\%                   \\
+bothinst+mock+reit+right+info+name+pos & \textbf{86.9} & \textbf{97}   & \textbf{91.7} & 81\%                  
\end{tabular}
\end{table}

The evaluation metrics (calculated against the \textsc{GRAD} label), as well as \textit{Template Stickiness}, for all the modifications detailed above are presented in Table~\ref{promptresults}. We provide these metrics rather than the more task-appropriate P@95\%R used above to make it easier to compare the various impacts of prompt changes. Any modification below 95\% Recall is presented in italic. \textit{Template Stickiness} refers to the percentage of outputs that fit a desired output format and contains the labels as defined in the prompt, meaning no further output parsing is necessary.

When multiple variants of a modification are evaluated, we either pick the best performing one or discard the modifications before applying the subsequent ones if there is no performance improvement.

We notice that the impact of prompt engineering on classification results is high. Simply asking the model to answer the question using its knowledge only reaches an F1-score of 65.6, with a Recall of 70.6, considerably short of our target, while our final prompt reaches an F1-score of \textbf{91.7} with \textbf{97}\% recall.

Interestingly, few-shot CoT prompting the model with examples performs noticeably worse than a zero-shot approach. We speculate that this is due to the examples biasing the model reasoning too much while the knowledge it already contains is sufficient for most classifications. Any attempt at providing more thorough reasoning for either label resulted in increased recall and decreased precision for the label. Despite multiple attempts, we found no scenario where providing examples performed better than zero-shot classification.

Providing instructions to the model, with a role description as its system message and an initial user message describing the text, yielded the single biggest increase in performance (+5.9F1). Additionally, we highlight the impact of small changes to guide the model's reasoning. Mocking an acknowledgement of the instruction allows the model to hit the 95\% Recall threshold (+1.3F1). Small additions, such as naming the model or providing it with positive reinforcement upon its acknowledgement of the instructions, also resulted in increased performance.

We found that \textsc{gpt-3.5.turbo} struggles with \textit{Template Stickiness}, which we did not observe with \textsc{text-davinci-003}. Its answers often required additional parsing, as it would frequently discard the (A)/(B) answering format asked of it. Requesting that it follows either a strict reasoning template or a loose answer template yielded considerably higher \textit{template stickiness} but resulted in performance decreases, no matter the template wording.

Overall, we find that these results highlight just how prompt-sensitive downstream results are, and we showcase a good overview of common techniques that can result in large performance improvements.

A limitation of this study is that we showcase the impact of prompt engineering and of various prompt modifications. However, we are unable to provide a fully reliable explanation as to why these modifications have such an impact. Large Language Models are trained on vast quantities of text to predict the next token, which then results in a quantity of emergent abilities \cite{emergent}, which can be elicited through specific prompting. While this prompting can intuitively make sense, there's a lack of theory as to how certain changes, such as adding a name, can generate noticeable improvements. This is an open area of research which we hope to contribute to in the future.

\section{Conclusion}
In this work, we have presented the task of Graduate Job Classification, highlighting its importance. We have then evaluated a series of classifiers on a real-world dataset, attempting to find which approach allows for the best filtering of non-graduate jobs while still meeting a sufficiently high recall threshold to not remove a large amount of legitimate graduate jobs in our curation efforts. In doing so, we showcased that the best-performing approach on this task is the use of Large Language Models (LLMs), in particular OpenAI's \textsc{gpt-3.5-turbo}.

Using language models for downstream tasks requires a different paradigm, where time is not spent on fine-tuning the model itself but on improving the \textbf{prompt}, a natural language query. We present our evaluation of various prompt modifications and demonstrate the large improvement in performance that can be obtained by proper \textbf{prompt engineering} to allow the language model to leverage its vast amounts of amassed knowledge. We believe our work, presenting a real-world case study of the strong performance of LLMs on text classification tasks, provides good insight into prompt engineering and the specific prompt-tuning necessary to accomplish certain tasks. We provide our full results, and the resulting prompt is currently being used to filter thousands of jobs on a daily basis, to help support future applications in this area. We provide our full results, the resulting prompt being currently used to filter thousands of jobs on a daily basis, to help support future applications in this area.
\bibliographystyle{splncs04}
\bibliography{bib}
\end{document}